# Investigating KAN-Based Physics-Informed Neural Networks for EMI/EMC Simulations

Kun Qian and Mohamed Kheir

University of Southern Denmark (SDU)
Center for Industrial Electronics
Institute of Mechanical and Electrical Engineering
Alsion 2, DK-6400 Sønderborg, Denmark
{kqian,kheir}@sdu.dk

**Abstract.** The main objective of this paper is to investigate the feasibility of employing Physics-Informed Neural Networks (PINNs), in particular Kolmogorov–Arnold Networks (KANs), for facilitating Electromagnetic Interference (EMI) simulations. It introduces some common EM problem formulations and how they can be solved using AI-driven solutions instead of lengthy and complex full-wave numerical simulations. This research may open new horizons for green EMI simulation workflows with less energy consumption and feasible computational capacity.

**Keywords:** Electromagnetic Compatibility (EMC), EMI simulations, full-wave simulations, Physics-Informed neural networks.

## 1 Introduction and Problem Definition

Electronic devices nowadays suffer from different electromagnetic emission problems typically within the Printed-Circuit Boards (PCBs). This is due to the fact that a single PCB can employ different connectivity standards and therefore emissions and interferences increase. A cross-section of a PCB is shown in Fig. 1 where it consists of different metallization layers and through-hole vias. Most EMI issues originate from PCBs and radiate into free-space or couple into other parts of the circuit. Simulating such PCB is needed in the early-stage of product design and can be helpful in detecting any potential problems.

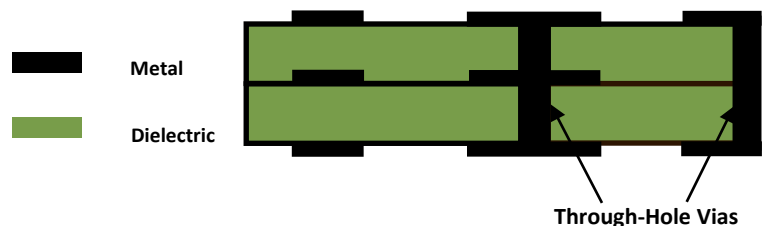

**Fig. 1**. A cross-section of a multi-layered PCB.

Traditional EMI simulation tools are mostly full-wave solvers where their simulation engines solve Maxwell's equations numerically to obtain the electric- and magnetic-field values within the structure. This also helps in determining the critical coupling paths and the propagation mechanism within the structure.

This problem can be solved either in time or frequency domain where the complexity increases dramatically as the frequency range increases. The well-known problems of EMI simulations include:

1. **The long simulation time:** A simulation may last hours or days depending on the size of the problem and the PC hardware we use. In simulations we need to cover a large space exploration and test all the possibilities.
2. **Computational capacity**: A typical EMC problem needs high-speed advanced CPU with high RAM capacity and additionally a number of GPU cards for simulation acceleration. This makes the simulation workstation very expensive.
3. **Energy consumption**: Long computation time on a highly-equipped machine consumes a huge amount of electricity and cooling.
4. **Simulation iterations**: Every modification to the EMI problems requires a new simulation to reach a solution. This is cumbersome since each iteration takes so long as declared in the first problem.

Data-driven methods, particularly those based on Machine Learning (ML), play a pivotal role in advancing the field of EMI/EMC. They enhance the understanding of coupling phenomena, guide design decisions, and improve system reliability [1,2]. For example, ML models can identify deviations from expected behavior, such as faulty components or abnormal radiation. ML can assist in designing and optimizing EMC systems, such as optimizing shielding materials [3] and grounding configurations [4].

On the other hand, the fusion of PINNs with conventional ML techniques has emerged as a transformative force. They seamlessly integrate domain-specific physics with ML [5]. For example, in an EMC setting, it can encode Maxwell's equations and fundamental principles, bridging theory and data. Moreover, PINNs is competitively data efficient thanks to its physics-based constraints, which can reduce dependence on computationally expensive simulations, accelerating EMI modeling. Such PINNs techniques typically require less datasets compared to traditional ML.

## 2   Methodology

PINNs have achieved remarkable success in solving Partial Differential Equations (PDEs) [6]. Multi-layer perceptrons (MLPs), which consist of neurons connected via trainable weights, are the standard building blocks for neural networks. Recent literature [7] has shown a promising alternative for MLPs: Kolmogorov–Arnold Networks (KANs), where neurons are connected via learnable activation functions. Fine-tuned

KANs can lead to small networks, which can further be described with symbolic functions, providing much better interpretability. Here, we present a feasibility study of applying physics-informed KANs to solve PDEs related to the physical laws of electromagnetics.

The problem under investigation is the simple 2D electrostatic box shown in Fig. 2 which can be described by the Laplace equation in (1).

$$-\nabla^2 u = 0, \tag{1}$$

where $u$ is the electric potential. The boundary conditions of the structure are illustrated in the same figure where the potential is zero everywhere except the excitation plane ($y = 1$).

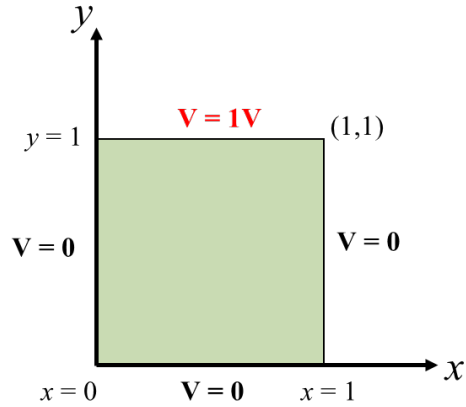

**Fig. 2** A 2D electrostatic box.

We use the KAN-based PINN to solve the above PDE problem, where the loss is defined as

$$\text{loss}_{\text{pde}} = \alpha \text{loss}_i + \text{loss}_b := \alpha \frac{1}{n_i}\sum_{i=1}^{n_i} |f(z_i)|^2 + \frac{1}{n_b}\sum_{i=1}^{n_b} |u(z_b)|^2, \tag{2}$$

where $\text{loss}_i$ and $\text{loss}_b$ denote the interior and boundary losses, respectively.

The interior loss is evaluated by a random sampling of $n_i$ points (denoted with $z_i$) inside the domain while the boundary loss is evaluated by a uniform sampling of $n_b$ points (denoted with $z_b$) on the boundary. $f(z_i)$ is defined as $f(z_i) = -\nabla^2 u(z_i)$, and $\alpha$ is the hyperparameter balancing the effect of the two loss terms.

Fig. 3 shows an example of KAN-based PINN methodology, where the neurons from different layers are connected with learnable activation functions ($u_{xx}$ and $u_{yy}$ denote the second-order derivative with respect to $x$ and $y$).

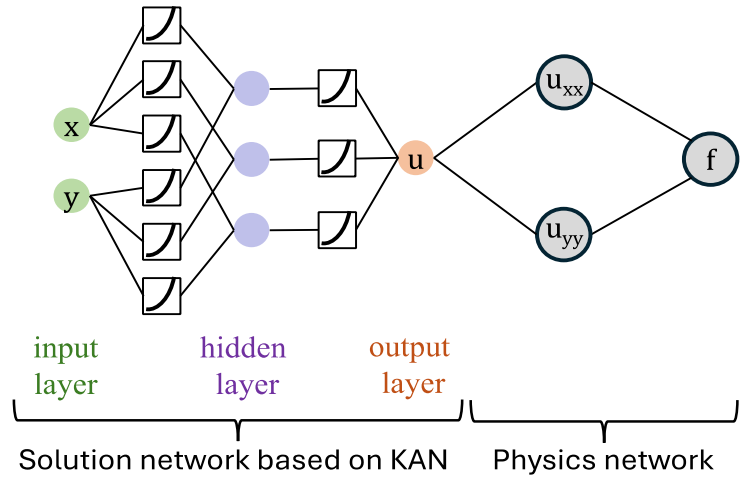

Fig. 3 Example PINN with KAN.

In comparison, Fig. 4 shows an example of MLP-based PINN methodology, where the neurons from different layers are connected with learnable weights.

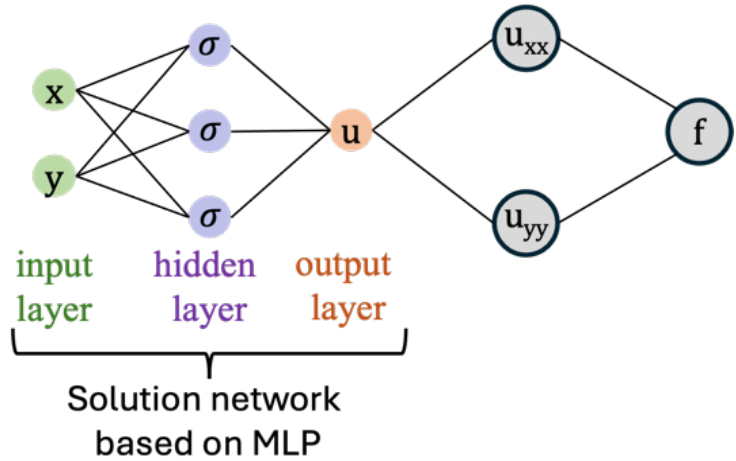

Fig. 4 Example PINN with MLP.

The input layer requires two neurons representing the coordinates, and the output of the neural network is comprised of one neuron representing the solution. Table 1 describes the two different types of PINNs used to solve the PDE.

**Table 1.** Network descriptions.

| NN Type | No. Hidden Layers | No. Neurons | Activation Function |
|---------|-------------------|-------------|---------------------|
| MLP     | 2                 | 32,32       | tanh                |
| KAN     | 2                 | 5,5         | *learnable*         |

For training, we uniformly select 50 points on each boundary side and randomly select 2500 points inside the domain. The trained models are used to predict the solution values for a $101 \times 101$ uniform grid, as provided by the simulation software.

## 3  Results

The electric potential of the electrostatic box is simulated in CST Studio Suite [8] and the result is shown in Fig. 5.

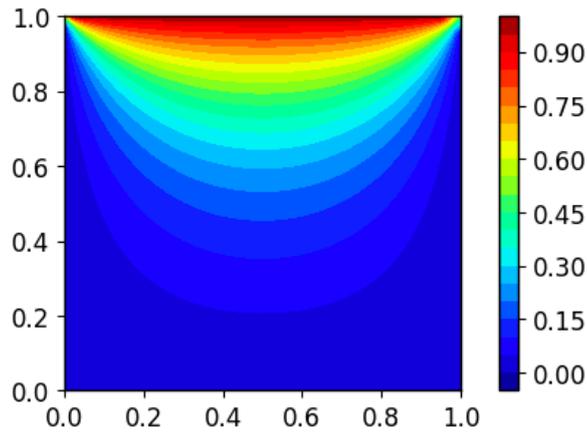

**Fig. 5** Ground truth solution provided by CST Studio Suite.

As clear from the figure, the distribution of the scalar potential is maximum at the center of the excitation plane and starts to decay at the boundary edges where the potential is zero elsewhere.

Fig. 6 and Fig. 7 show the results from the MLP-based and KAN-based PINNs, respectively. As shown, the two PINNs can achieve very similar results to those provided by the commercial simulation software.

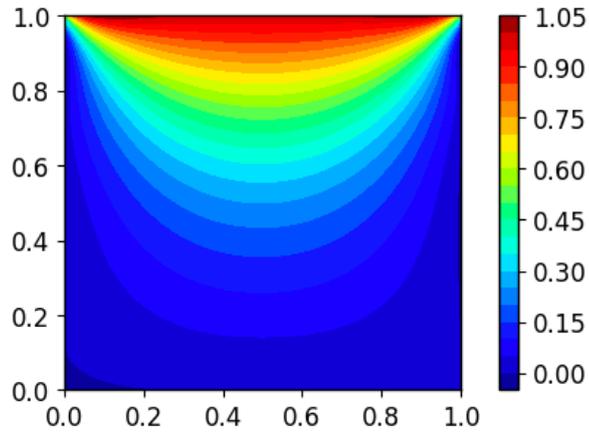

**Fig. 6** Results from the MLP-based PINN.

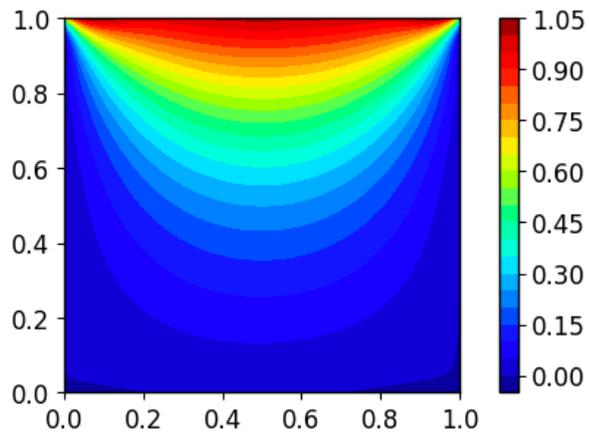

**Fig. 7** Results from the KAN-based PINN.

The slight (absolute) differences are shown in Fig. 8 and Fig. 9 where it did not exceed 0.1.

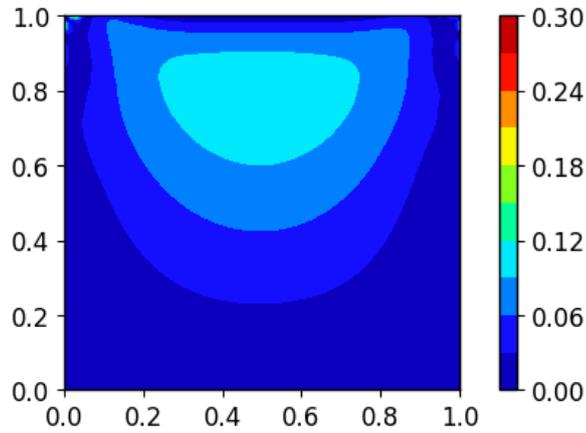

**Fig. 8** Comparing the absolute error between the results from the MLP-based PINN and the ground truth solution.

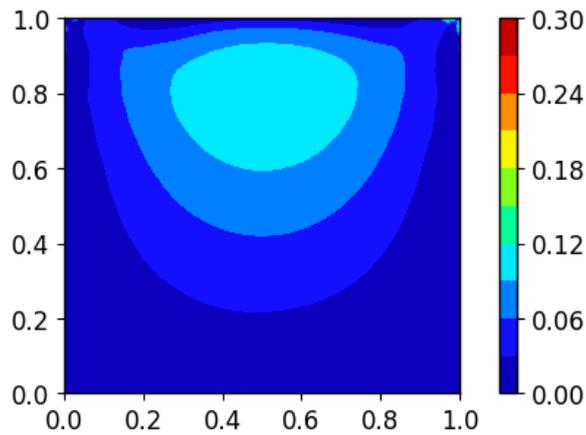

**Fig. 9** Comparing the absolute error between the results from the KAN-based PINN. and the ground truth solution.

## 4  Conclusions

This study has investigated the possibility of employing AI-driven techniques for EMI/EMC simulations as a green alternative. We have demonstrated the results from both MLP- as well as KAN-based PINNs where the KAN-based PINNs can achieve the same results as MLP-based PINN, but with much fewer neurons. This can consequently reduce the training time required and introduce an energy-efficient and fast simulation workflow compared to numerical full-wave simulations. This will introduce a promising potential solution for solving EMI/EMC problems and structures that normally require complex computational capacity, high energy consumption and a long time.